\documentclass{IEEEmce}

\usepackage{cite}
\usepackage{amsmath,amssymb,amsfonts}
\usepackage{graphicx}
\usepackage{textcomp}
\usepackage{wrapfig}
\usepackage{rotating}
\usepackage{mathtools}
\DeclarePairedDelimiter{\nint}\lfloor\rceil
\usepackage[T1]{fontenc}
\usepackage[retain-explicit-plus]{siunitx}
\usepackage{booktabs}
\usepackage{bigdelim}
\usepackage[ruled,vlined]{algorithm2e}
\usepackage{listings}
\usepackage[hyphens]{url}
\usepackage{lineno,hyperref}
\usepackage{pifont}
\newcommand{\cmark}{\ding{51}}
\newcommand{\xmark}{\ding{55}}
\usepackage{nameref}
\usepackage{algorithmicx}
\usepackage{algpseudocode}

\usepackage{blindtext}
\usepackage{hyperref}
\usepackage{nameref}

\newcounter{mylabelcounter}

\makeatletter
\newcommand{\labelText}[2]{%
\refstepcounter{mylabelcounter}%
\immediate\write\@auxout{%
 \string\newlabel{#2}{{\unexpanded{#1}}{\thepage}{{\unexpanded{#1}}}{mylabelcounter.\number\value{mylabelcounter}}{}}%
}%
}
\makeatother

\definecolor{new_blue}{RGB}{40, 185, 242}

\jname{IEEE Consumer Electronics Magazine}
\pubyear{2023}

\begin{document}

\twocolumn[{
\textcopyright 2024 IEEE. Personal use of this material is permitted. Permission from IEEE must be obtained for all other uses, in any current or future media, including reprinting/republishing this material for advertising or promotional purposes, creating new collective works, for resale or redistribution to servers or lists, or reuse of any copyrighted component of this work in other works. DOI: \url{https://doi.org/10.1109/MCE.2024.3387019}
}]

\title{Unsupervised explainable activity prediction in competitive Nordic Walking from experimental data}

\author{Silvia García-Méndez}
\author{Francisco de Arriba-Pérez}
\author{Francisco J. González-Castaño}
\affil{Information Technologies Group, atlanTTic, University of Vigo}

\author{Javier Vales-Alonso}
\affil{Communication and Information Technologies Department, Technical University of Cartagena}

\begin{abstract}
Artificial Intelligence (\textsc{ai}) has found application in Human Activity Recognition (\textsc{har}) in competitive sports. To date, most Machine Learning (\textsc{ml}) approaches for \textsc{har} have relied on offline (batch) training, imposing higher computational and tagging burdens compared to online processing unsupervised approaches. Additionally, the decisions behind traditional \textsc{ml} predictors are opaque and require human interpretation. In this work, we apply an online processing unsupervised clustering approach based on low-cost wearable Inertial Measurement Units (\textsc{imu}s). The outcomes generated by the system allow for the automatic expansion of limited tagging available (\textit{e.g.}, by referees) within those clusters, producing pertinent information for the explainable classification stage. Specifically, our work focuses on achieving automatic explainability for predictions related to athletes' activities, distinguishing between correct, incorrect, and cheating practices in Nordic Walking. The proposed solution achieved performance metrics of close to \SI{100}{\percent} on average.
\end{abstract}

%\begin{IEEEkeywords}
%Clustering, Interpretability and Explainability, Human Activity Recognition, Nordic Walking, Supervised Machine Learning, Wearable sensors.
%\end{IEEEkeywords}

\maketitle

\section*{INTRODUCTION}

Automatic Human Activity Recognition (\textsc{har}) \cite{Jobanputra2019,Al-Hammadi2020} is a field of great interest for sports research. \textsc{har} is based on the premise that body movements produce differentiated patterns of signals that can be collected with sensors such as visual recognition and Inertial Measurement Units (\textsc{imu}s).

Sensors in the literature on sport \textsc{har} can be broadly divided into three groups: (\textit{i}) portable devices or wearables (\textit{e.g.}, gyroscopes and accelerometers), (\textit{ii}) environmental sensors (\textit{e.g.}, cameras and \textsc{gps}), and (\textit{iii}) sensors integrated into personal terminals (\textit{e.g.}, smartphones). Integrated sensors are popular and allow for app-based business models, but wearable sensors are preferred for stringent scenarios that require freedom of movement \cite{Ayman2019,Abbaspour2020,Bozkurt2022}, as it is often the case with sports practice \cite{Navalta2019}.

Current methodologies to evaluate sports performance involve predictive techniques based on statistical methods. Performance in this setting, however, is the joint result of a wide variety of factors, including level of training, physical condition, and team interactions \cite{Mohamad2018}. More complex, intelligent, methodologies are therefore needed to characterize, beyond performance, specific states during the practice of sport or to predict certain events that may lead to injuries \cite{Challa2021}. 

These emerging needs have led to the application of Artificial Intelligence (\textsc{ai}) such as Machine Learning (\textsc{ml}) \cite{Cust2019}. Nevertheless, decisions made by traditional \textsc{ml} predictors are opaque and require human interpretation of the reasons underlying the decisions. This work focuses on explainable \textsc{ai} (\textsc{xai}), a relatively unexplored field in the area of sport. \textsc{xai} techniques are designed to infer the reasons behind prediction outcomes\footnote{Available at \url{https://doi.org/10.48550/arXiv.2302.05624}, March 2024.}. In this study, we leverage \textsc{xai} to provide automatic interpretations and explanations of the differentiation between correct, incorrect, and cheating practices in a Nordic Walking case study using low-cost \textsc{imu}s. On a more detailed level, we analyze and leverage the most relevant features of the \textsc{ml} model to improve explainability.

The rest of this paper is organized as follows. The section \textsc{related work} reviews relevant work on \textsc{har} in sports using supervised and unsupervised \textsc{ml} techniques. The section \textsc{methodology} describes the solution for explainable unsupervised characterization of Nordic Walking practice. The section \textsc{experimental results} presents the experimental data and implementations used and the results obtained. Finally, the section \textsc{conclusions} provides overall discussion and highlights the future scope of research and study.

\section*{RELATED WORK}
\label{sec:related_work}

Numerous research works in the field of sport have analyzed action and movement \cite{Cust2019} as well as fitness and performance \cite{Borms2018}. Many of these studies use low-cost commercial \textsc{imu}s, such as accelerometers, gyroscopes, and magnetometers. Examples of outputs measured include basketball movements \cite{Hu2020}, tennis and table tennis strokes \cite{Tabrizi2021}, and volleyball training \cite{Vales-Alonso2015}.

The more recent \textsc{ml} solutions employ big-data and small-data models. Big-data models are typically deep learning models due to the large volumes of data involved \cite{Lerebourg2022,Bai2019}. However, data volumes in sports training scenarios are generally small, corresponding to a single individual and even a single training session. Unfortunately, supervised approaches have dominated in \textsc{har} \cite{Shen2020}. 

Supervised methodologies for the analysis of sports practice include Decision Trees (\textsc{dt}) \cite{Bulac2016}, Support Vector Machines (\textsc{svm}) \cite{Pisner2020}, \textit{k}-nearest neighbors (\textit{k}-\textsc{nn}) \cite{Cunningham2022}, and ensemble methods such as boosting, bagging, and stacking \cite{Pavlyshenko2018}. Notably, Noor \textit{et al.} (2017) \cite{Noor2017} differentiated between transitional and non-transitional activities. In contrast, Bulbul \textit{et al.} (2018) \cite{Bulbul2018} recognized six activities: walking, climbing up and down stairs, sitting, standing, and lying down. Zainudin \textit{et al.} (2018) \cite{Zainudin2018} combined one-versus-all (\textsc{ova}) models with a self-adaptive algorithm to select features for sports practice analysis. In addition, Bharti \textit{et al.} (2019) \cite{Bharti2019} studied different body sensor locations on the extremities for activity recognition. Gil-Martin \textit{et al.} (2020) \cite{Gil-Martin2020} applied deep learning models to annotated data to identify movements during activity, while Zhu \textit{et al.} (2020) \cite{Zhu2020} predicted athlete performance using an \textsc{svm} model. Adopting a different approach, Rossi \textit{et al.} (2021) \cite{Rossi2021} studied injury prevention during training, and Webber \& Rojas (2021) \cite{Webber2021} fused data from accelerometer and gyroscope sensors at sensor, feature, and decision levels to apply diverse well-known \textsc{ml} models (\textsc{dt}, \textit{k}-\textsc{nn}, Linear Discriminant Analysis (\textsc{lda}) and \textsc{svm}). Although the best results were obtained with decision-level fusion, the authors concluded that the computational power and processing times required were unacceptable for a practical \textsc{har} system. Zeng \textit{et al.} (2021) \cite{Zeng2021} predicted effort levels during the execution of physical activities using acceleration and angular velocity sensors on the arms, waist, and wrists. Finally, Zhang \& Li (2022) \cite{Zhang2022} used neural networks and \textsc{svm}s to analyze athlete movements and the impact of training equipment. 

Unsupervised \textsc{ml} techniques \cite{Suto2018}, which, when applicable, are highly convenient as they do not require manual tagging and can be used to detect practice patterns. However, \textit{a priori}, a specific pattern may be complex to assign to a specific state (\textit{i.e.}, sports execution phase). There are few works on unsupervised \textsc{ml} techniques in sports. Domingo \textit{et al.} (2018) \cite{Domingo2018} applied Subsequence Time Series (\textsc{sts}) analysis of dominant arm acceleration data using \textit{K}-means clustering \cite{Sinaga2020}, while Gu \textit{et al.} (2018) \cite{Gu2018} proposed temporarily grouping large volumes of unlabeled data into duly annotated time points. Conversely, Van Kuppevelt \textit{et al.} (2019) \cite{VanKuppevelt2019} applied an unsupervised Hidden Semi-Markov Model to automatically segment and cluster data from wearable acceleration sensors to infer the type of activity performed from the clusters detected. Finally, Janarthanan \textit{et al.} (2020) \cite{Janarthanan2020} employed an unsupervised deep learning-assisted reconstructed encoder for sports activity recognition, while Serantoni \textit{et al.} (2022) \cite{Serantoni2022} used a \textit{K}-means model to differentiate between overexertion stages in cardiovascular exercise.

Nordic Walking involves complex interactions between various body parts. When performed incorrectly, it increase the risk of injury and reduces training effectiveness. A typical observable error in Nordic Walking, for example, is dragging the tips of the poles over the ground \cite{Derungs2018}. To prevent cheating, in Nordic Walking competitions, judges positioned along the track evaluate whether participants are purposely using incorrect techniques, such as when running instead of walking. Some works on automatic activity monitoring based on integrated pole sensors, heart sensors, and \textsc{gps} data \cite{Mocera2018,Pierleoni2022} have used static formulae, that is, formulae not involving multivariate or intelligent data analysis such as \textsc{ml}. One clear exception is the work by Derungs \textit{et al.} (2018) \cite{Derungs2018}, which proposed a supervised regression system to detect incorrect practices based on \textsc{imu}s placed in seven fixed body locations. Wiktorski \textit{et al.} (2019) \cite{Wiktorski2019} employed Dynamic Time Warping (\textsc{dtw}) and a semi-supervised approach to differentiate between Nordic Walking and other activities. This is the only example to our knowledge of the application of a, to an extent, unsupervised technique to this sport.

Finally, and particularly relevant to the scope of this work, it should be noted that \textsc{xai} \cite{Kumar2023}, which has been applied to a wide range of applications \cite{DeArriba-Perez2022,Gonzalez-Gonzalez2022}, is largely missing in elite sports \cite{Ehatisham-ul-Haq2020}, with a few exceptions based on strictly offline (batch) supervised solutions. Examples include explainable game-play prediction \cite{Wang2022} and case-based reasoning to recommend training plans to marathon runners using a visual dashboard \cite{Feely2020}. Sun \textit{et al.} (2020) \cite{Sun2020} studied the trade-off between accuracy and transparency in sports analytics. They employed a tree version of a neural network to generate visual (not natural language) descriptions of ice hockey and soccer practice predictions. Finally, Lisca \textit{et al.} (2021) \cite{Lisca2021} applied an extreme gradient boosting ensemble model in batch mode to obtain symbolic insights into goalkeeper kinematics from a single motion sensor.

\subsection*{RESEARCH CONTRIBUTION}

To the best of our knowledge, the solution is the first work in sport \textsc{har} that applies explainability techniques to infer knowledge from unsupervised small-data analysis in online processing mode, taking Nordic Walking as a case study. Unlike batch processing, online processing updates model profiling and classification with each incoming sample \cite{GarciaMendez2022}. Therefore, the system can be easily deployed in the field without imposing a manual burden on operators. 

A generalistic supervised \textsc{har} system must be trained in advance. This is a huge, costly, and laborious undertaking involving many practitioners. That said, there is a finite known set of well-defined practice patterns in certain sports, both in training and competition scenarios. This is the case with Nordic Walking, where these well-defined patterns include correct practice, cheating, and, in the case of coaching, incorrect practices. We demonstrate that it is possible to cluster these patterns very efficiently, to a level comparable to that seen with supervised systems. Then, with minimal burden, it is possible to label just some data samples within each cluster. This would be a quite natural process in Nordic Walking competitions. Judges could easily tag these references during a race by clicking a button when a practitioner passes by their locations. Finally, once the labels are propagated within their respective clusters, the data could be re-classified for explainability. It should be noted that {\it the idea of this re-classification is not to improve classification performance} (since the clustering stage has already classified the data samples by propagating reference labels) but to apply a supervised tree classification method that, unlike clustering, is {\it intrinsically explainable}. Accordingly, we ultimately achieve classification at no tagging cost and produce useful explainability information.

The base clustering technique, which generates the labels for the explainability classification stage by expanding the limited {\it a posteriori} tagging available within the clusters, achieves up to \SI{97.68}{\percent} accuracy. Supervised learning is only applied in this research for two purposes: (\textit{i}) as previously explained, to extract the relevant features for the explainability module with minimal labor, and (\textit{ii}) as a baseline for the comparisons in the experimental evaluation. Summing up, unsupervised clustering enables automatic tagging expansion, a solution to a prevalent challenge in commercial \textsc{har} \cite{Bi2021}. This feature is subsequently leveraged to train an automatically supervised classifier for explainability purposes at minimal cost.

\textsc{har} is of great interest to consumer electronics in sports and health \cite{Fu2020,Meng2020,Herencsar2023}. Some representative examples of commercial products are Fibaro\footnote{Available at \url{https://www.fibaro.com}, March 2024.}, Fitbit\footnote{Available at \url{https://www.fitbit.com}, March 2024.}, and Mi Band\footnote{Available at \url{https://www.mi.com/global/miband}, March 2024.}. The target consumers of a commercial version of our system, rather than individual practitioners, would be group coaches, course organizers, small event organizers, and similar. The implementation could be based on low-cost consumer electronics using smartphones and compact accelerometers as described in the \textsc{Experimental Results} section.

Table \ref{tab:comparison} provides a brief comparison of our work with most of the related previous works discussed in this section.

\begin{table*}[!htbp]
\centering
\caption{\label{tab:comparison}Comparison with previous research on intelligent analysis of Nordic Walking practice.}
\begin{tabular}{lccc}
\toprule
\multirow{1}{*}{\textbf{Authorship}} & \textbf{Approach} & \textbf{Stream (online) processing} & \textbf{Explainability} \\
\midrule

Mocera \textit{et al.} (2018) \cite{Mocera2018} & \multirow{2}{*}{Rules \& formulas} & \xmark & \xmark\\

Pierleoni \textit{et al.} (2022) \cite{Pierleoni2022} & & \xmark & \xmark\\\midrule

Derungs \textit{et al.} (2018) \cite{Derungs2018} & Supervised \textsc{ml} & \xmark & \xmark\\\midrule

Wiktorski \textit{et al.} (2019) \cite{Wiktorski2019} & Semi-supervised \textsc{ml} & \xmark & \xmark\\\midrule\midrule

\textbf{Our solution} & Unsupervised \textsc{ml} & \cmark & \cmark\\
\bottomrule
\end{tabular}
\end{table*}

\section*{METHODOLOGY}
\label{sec:proposed_method}

\begin{figure*}[!htbp]
\centering
\includegraphics[scale=0.15]{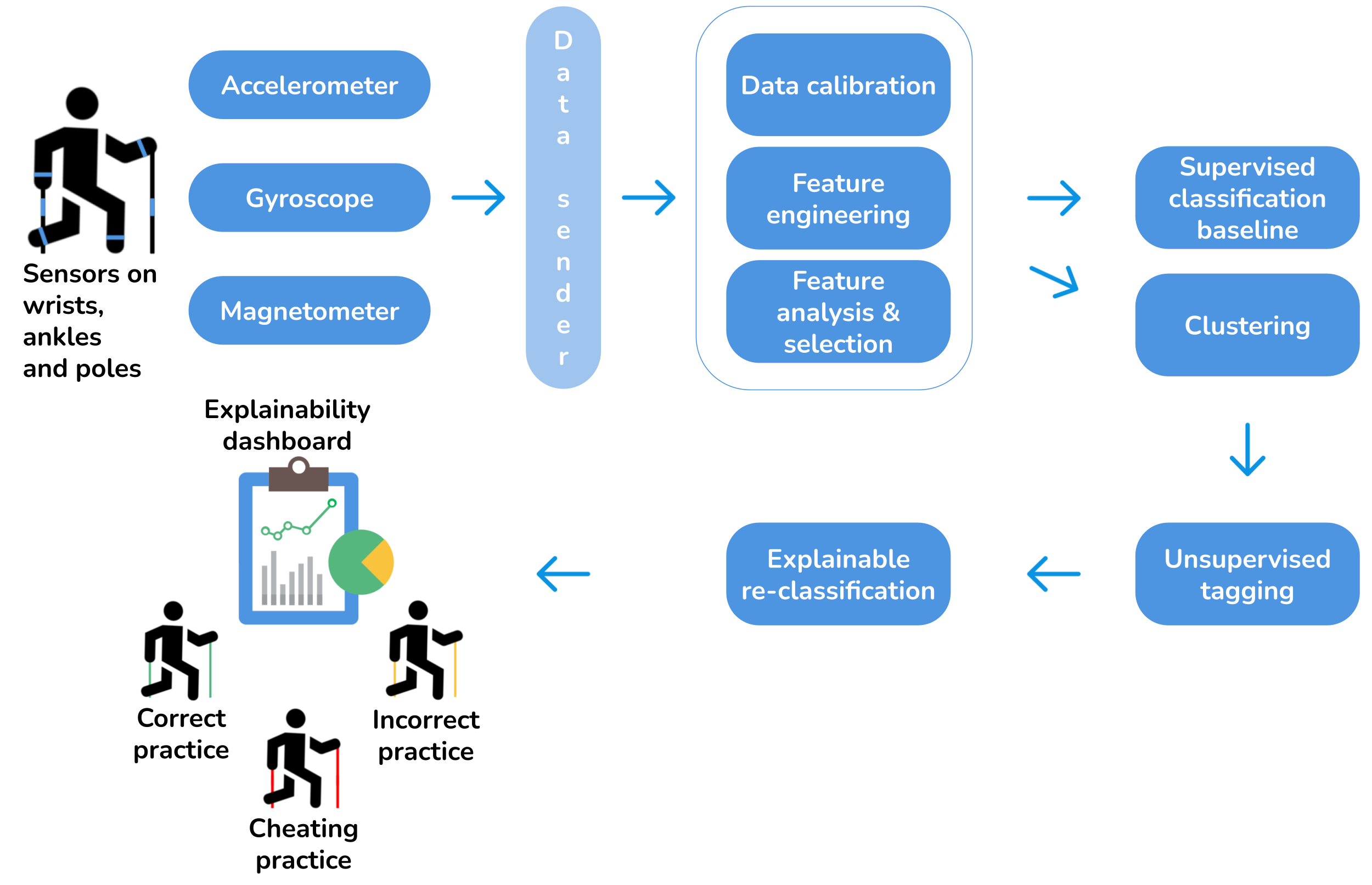}
\caption{\label{fig:scheme}Nordic Walking practice assessment scheme.}
\end{figure*}

Figure \ref{fig:scheme} shows the scheme of the solution. It comprises the \textsc{imu}s (with accelerometers, gyroscopes, and magnetometers) placed on wrists, ankles, and poles; a data processing module composed of data calibration, feature engineering and analysis \& selection stages; and an online processing clustering module, whose outcome is compared with a supervised classification baseline. Moreover, the automatically expanded labels from the unsupervised clustering module train a supervised classifier for explainability purposes. Finally, the explainability module provides visual and natural language descriptions of the prediction outcome (\textit{i.e.}, correct, cheating, or incorrect practices).

\subsection*{WEARABLE SENSORS}
\label{sec:sensors}

\begin{figure*}[!htbp]
\centering
\includegraphics[width=0.8\textwidth]{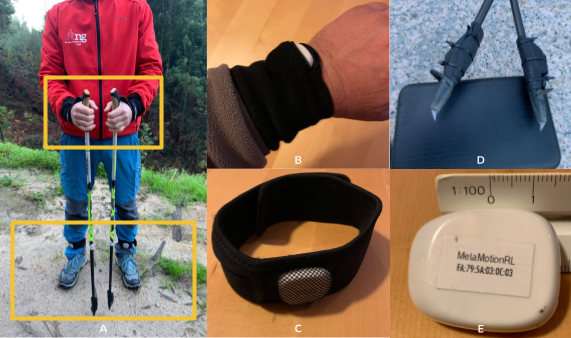}
\caption{\label{fig:sensors_position}Arrangement of the \textsc{imu}s. \textsc{a}) positions on body and poles, \textsc{b}) wrist \textsc{imu}, \textsc{c}) ankle \textsc{imu}, \textsc{d}) pole \textsc{imu}s, \textsc{e}) \textsc{imu} size in cm.}
\end{figure*}

Three pairs of wearable \textsc{imu}s are placed on the left and right sides of the body ($P = \{left,right\}$), specifically, on wrists, ankles, and poles ($L = \{wrist,ankle,pole\}$). Each \textsc{imu} includes tri-axial ($A = \{x,y,z\}$) accelerometers (to capture the modulus and direction of the movement accelerations), gyroscopes (to capture the axis rotation speeds), and magnetometers (to capture the modulus and direction of the magnetic fields) ($S = \{accelerometer,gyroscope,magnetometer\}$), placed as shown in Figure \ref{fig:sensors_position}. These \textsc{imu}s gather movement data from the athletes. According to Edwards \textit{et al.} (2019) \cite{Edwards2019}, the maximum average acceleration in human walking is \num{4.5} g, so the operation range of the wearable sensors must cover this with a reasonable excess margin without compromising sensitivity.

The real-time data from the sensors is used as the input data for the data processing, online processing \textsc{ml} clustering, and explainability modules. Of interest to consumer electronics is that, even though the sensors are positioned in specific locations on the poles and the body, precise placement relative to body parts is unnecessary. Our tests showed significant linear and angular displacements of the sensors between and during sessions, even within the same user. No special care was taken when strapping sensors into place during the data-gathering process.

\subsection*{DATA CALIBRATION, FEATURE ENGINEERING, ANALYSIS \& SELECTION}
\label{sec:data_processing}

During the data processing stage, raw data gathered by the wearable sensors are merged and filtered to ensure high-quality classification and explainability results. There are two fundamental data processing stages: first data calibration, and then, online processing feature engineering and analysis \& selection.

\begin{description}

\item \textbf{Data calibration}. Noise is a major issue when working with analogical signals. \textsc{har} signals are typically filtered using Finite Impulse Response (\textsc{fir}) filters \cite{Slim2019,Stuart2021}. In this work, we apply an averaging \textsc{fir} filter and produce other features using four sliding windows of different sizes.

Different window lengths are set at the calibration stage by inspecting the sample intervals between successive minima of the sensor signals. To guarantee that at least two minima are captured per sensor, and given the repetitive limb and pole movements in Nordic Walking, the calibration stage lasts for 90 seconds. Let $k$ be the size of the calibration stage in samples. For each sensor signal $x_{p,a,s,l}$, we define $v_{p,a,s,l}$ as follows (we will avoid the $p,a,s,l$ sub-indices for clarity unless strictly necessary to explain the formulae): \\

\noindent 
Let $v_1$, $v_2$ be sample indexes in $[1,k-1]$, and $V_{aux}$ a subset of sample indexes, such that \eqref{eq:min}:

\begin{equation}\label{eq:min}
\begin{split}
V_{aux} = \{ n\in[1,k-1] \mid \\
x[n-1] > x[n] < x[n+1]\} \\
v_1= min(V_{aux}) \\
v_2= min(V_{aux} \backslash \{v_1\}) \\
v=v_2-v_1
\end{split}
\end{equation}

That is, $v_{p,a,s,l}$ is approximately the number of samples between the first two local minima delimiting an $x_{p,a,s,l}$ activity interval in the calibration stage. Let $V=\{v_{p,a,s,l}\}$, $|V|=|P|\times|A|\times|S|\times|L|$, and let $V' = \{v'_0,\ldots,v'_{|V|-1}\}$, $v'_0 \leq v'_1 \leq \ldots \leq v'_{|V|-1}$ be a sequence of reordered indexes of $V$, that is, $\forall v \in V,$ $v \in V'$, in increasing order. Then, the four window lengths $w_{Q1}$, $w_{Q2}$, $w_{Q3}$ and $w_{avg}$ for all sensors are defined as follows \eqref{eq:statistics}, where $r_{min}$ and $r_{max}$ are the minimum and maximum data rates for all sensors used:

\begin{equation}\label{eq:statistics}
\begin{split}
w_{Q1}=\nint{2\frac{r_{max}}{r_{min}}}v'_{\nint{\frac{1}{4}|V|}}\\
w_{Q2}=\nint{2\frac{r_{max}}{r_{min}}}v'_{\nint{\frac{2}{4}|V|}}\\
w_{Q3}=\nint{2\frac{r_{max}}{r_{min}}}v'_{\nint{\frac{3}{4}|V|}}\\
w_{avg}=\nint{2\frac{r_{max}}{r_{min}}}\nint{\frac{1}{|V|}\sum_{i=0}^{|V|} v'_i}
\end{split}
\end{equation}

\item \textbf{Feature engineering}. Once the sliding windows are selected, online processed features can be calculated $\forall n > max(w_{Q1}, w_{Q2}, w_{Q3}, w_{avg})$ (to avoid a cold start) for each sensor $x_{p,a,s,l}$ and window size $w$: average (${avg}^w_{p,a,s,l}$), standard deviation (${std}^w_{p,a,s,l}$), quartile values (${Q}^w_{1,p,a,s,l}$- ${Q}^w_{3,p,a,s,l}$) and maximum modulus of the components of the Fast Fourier Transform (\textsc{fft}) ($F^w_{p,a,s,l}$), as follows \eqref{eq:feature_engineering}:

\begin{equation}\label{eq:feature_engineering}
\begin{split}
\forall w \in \{w_{Q1},w_{Q2}, w_{Q3},w_{avg}\}, \; \forall n\geq \{w,k\} \\ \\
X[n] = \{ x[n-w+1],\ldots,x[n]\}. \\
Y[n] = \{y_0[n], y_1[n],\ldots,y_{w-1}[n]\} \mid \\ y_0[n]\leq y_1[n]\leq\ldots\leq y_{w-1}[n], \\
\mbox{where} \; \forall x \in X[n], \; x \in Y[n]. \\ \\
Q^w_{1}[n]=y_{\nint{\frac{1}{4}w}}[n] \\
Q^w_{2}[n]=y_{\nint{\frac{2}{4}w}} [n]\\
Q^w_{3}[n]=y_{\nint{\frac{3}{4}w}} [n]\\
avg^w[n]=\frac{1}{w}\sum_{i=0}^{w} y_i [n]\\
std^w[n]=\sigma(X[n]) \\
F^w[n]=|FFT(X[n])|_\infty
\end{split}
\end{equation}

\item \textbf{Feature analysis \& selection}. 

 Let $t_\sigma>0$ be a configurable threshold, and $\Phi$ a transformation operator such that $\Phi(\{x\})=\{x\}$ if $x>t_\sigma$, $\Phi(\{x\})=\emptyset$ otherwise. At any slot $n$, $n\geq k$, by considering the online processed standard deviations of all possible features at that moment, the set of selected features $S^{w}_{p,a,s,l}[n]$ for online processing prediction and training update at that slot is composed of $\Phi(\sigma(\{{avg}^w_{p,a,s,l}[k],\ldots,$ ${avg}^w_{p,a,s,l}[n]\}))$, $\Phi(\sigma(\{{std}^w_{p,a,s,l}[k],\ldots,$ ${std}^w_{p,a,s,l}[n]\}))$, $\Phi(\sigma({Q}^w_{i,p,a,s,l}[k],\ldots,$ ${Q}^w_{i,p,a,s,l}[n]))$, $i=1,2,3$, and $\Phi(\sigma(\{{F}^w_{p,a,s,l}[k],\ldots,$ ${F}^w_{p,a,s,l}[n]))$.

\end{description}

\subsection*{ONLINE PROCESSING}
\label{sec:classification}

Incremental profiling is performed at each slot $n$, using the set of features $S^{w}_{p,a,s,l}[n]$. Incremental classification is composed of a number of steps, described below.

\paragraph*{ONLINE CLUSTERING AND UNSUPERVISED TAGGING}

Unsupervised clustering is based on the \textit{K}‐means method \cite{Sinaga2020}. The results are evaluated using the best possible class mapping, that is, by checking the possible matchings between each cluster discovered and the target categories and selecting the matching that maximizes classification success. In practice, it can be assumed \textit{a posteriori} that the samples taken near the (few) judges present during a Nordic Walking competition are correctly tagged. In this work, this tagging expansion leads to three class labels, $c_0$, $c_1$, and $c_2$, corresponding respectively to correct practice, cheating, and incorrect practice.

\paragraph*{ONLINE PROCESSING SUPERVISED CLASSIFICATION BASELINE}

As a baseline, the following \textsc{ml} algorithms are applied, based on their good performance in \textsc{har} problems in the literature \cite{Shen2020,Khannouz2020,Chen2022}.

\begin{itemize}
 
 \item \textbf{Gaussian Naive Bayes} (\textsc{gnb}) \cite{Xue2021}, based on a Gaussian distribution, for stream-based classification with the traditional Naive Bayes (\textsc{nb}) model.
 
 \item \textbf{Hoeffding Adaptive Tree Classifier} (\textsc{hatc}) \cite{Stirling2018}, an online processing single tree-based model with a branch performance monitoring mechanism.
 
 \item \textbf{Adaptive Random Forest Classifier} (\textsc{arfc}) \cite{Gomes2017}, a stream-based tree ensemble majority voting with re-sampling and random feature selection using the concept drift mechanism.
 
\end{itemize}

The accuracy, precision (micro and macro), recall (micro and macro), and elapsed time of these algorithms are calculated with the predictive sequential (\textit{i.e.}, prequential evaluation) protocol for online processing learning \cite{Gama2013}.

\paragraph*{EXPLAINABLE UNSUPERVISED RE-CLASSIFICATION} 
\label{sec:explainability}

The explainability module traverses the estimator decision path of an \textsc{arfc} model (also used as one of the supervised classification baseline algorithms) to extract the components of features $S^{w}_{p,a,s,l}[n]$ $\forall w, p, a, s, l$, $\forall n > n_{init}=max(w_{Q1},$ $w_{Q2},$ $w_{Q3},$ $w_{avg})$, which are all relevant for explainability purposes. The reclassification algorithm in this module is trained using tags $c_0$, $c_1$, and $c_2$ resulting from the clustering stage, which are translated to the class references available by expanding these references inside the corresponding clusters. Therefore, the clustering stage is explained from an interpretable \textsc{ml} algorithm with automatic unsupervised tagging because the tags are not manually produced. Let $S_{i,p,a,s,l}^w[n]$ be the $i$-th feature component, $i=1...6$, of the 6-dimensional vector $S_{p,a,s,l}^w[n]$ defined from the metrics $Q^w_{1}[n], Q^w_{2}[n], Q^w_{3}[n], avg^w[n], std^w[n]$ and $F^w[n]$ in \eqref{eq:feature_engineering}, in this same order. Note that, at slot $n$, a component will be only defined if its updated standard deviation exceeds threshold $t_\sigma$ (for example, $S_{1,p,a,s,l}^w[n]=Q^w_{1,p,a,s,l}[n]$ when $Q^w_{1,p,a,s,l}[n]$ > $t_\sigma$). Otherwise, it will be left undefined. In the sequel, undefined components are ignored in the calculations. Once the \textsc{arfc} trees are obtained for the current prediction, all the paths leading to the corresponding class are traversed. Frequency $\xi^{w}_{i,p,a,s,l}[n]$ of subset $\{S_{i,p,a,s,l}^w[n], n>=n_{init}\}$ is the number of times the tuple $(i,w,p,a,s,l)$ is used as an index in those paths, as shown in Algorithm \ref{alg:relevant_features}. The default feature component selected to display on the visual dashboard is $S^{w^*}_{i^*,p^*,a^*,s^*,l^*}[n]$, $(i^*,w^*, a^*, p^*, s^*, l^*)=argmax_{(i,w, a, p, s, l)}(\xi^{w}_{i,p,a,s,l}[n])$, which becomes the first element of the ordered list $\Gamma[n]$. The rest of the elements in this list are components $\{S_{i,p,a,s,l}^w[n], n>n_{init}\}$ ordered in decreasing order of the frequencies taken from $\xi^{w}_{i,p,a,s,l}[n]$.

\begin{algorithm*}[htb]\caption{Extraction of relevant features}
\label{alg:relevant_features}
 \SetAlgoLined
 \KwData{ $classifier$, $prediction$, $sample$ \tcp*[h]{Output of the classification model.}\\}
 \KwResult{Relevant feature components sorted by frequency. }
$feature\_list=[]$\\
\For{$estimator$ in $classifier$}{
\If{$prediction = estimator.predict(sample)$}{
 $node=estimator[0]$ \tcp*[h]{Root node. Node structure: \{feature,threshold,left\_branch,right\_branch,is\_terminal\_node\}.}\\
 
 \While{$node[is\_terminal\_node] \neq True$}{
 $feature= node[feature]$\\
 $threshold = node[threshold]$\\
 \eIf{$sample[feature] \leq threshold$}{
 $node=node[left\_branch]$
 }{
 $feature\_list.append(feature)$\\
 $node=node[right\_branch]$\\
 }
 }
 }
 }
 $\Gamma[n] = feature\_components\_sorted\_by\_frequency(feature\_list)$ \\
 return $\Gamma[n]$\\
\end{algorithm*}

Visual content (feature values from six body sensors, further divided into three axes and four sliding windows) and textual descriptions (generated in natural language) of athlete performance are shown, together with predictions of practice type (correct, incorrect, and cheating). 

\section*{EXPERIMENTAL RESULTS}
\label{sec:experimental_results}

This section describes the experimental data and wearable sensors used. Then, it outlines the different sub-problems in the tests, and finally, it presents and discusses the implementations and results observed for data calibration, feature engineering, analysis \& selection, online processing baseline and unsupervised classification, and unsupervised explainability.

\subsection*{EXPERIMENTAL DATA}
\label{sec:experimental_data}

The \textsc{nwgti} and \textsc{pamap2} data sets were used to analyze the system's performance. \textsc{nwgti} data were collected from training sessions involving five Nordic walkers under the guidance of Ignacio Garc\'ia P\'erez, regional Nordic Walking coach of the Galician Mountaineering Federation, Spain. His expertise in technique and competition refereeing enabled the identification of correct practices and typical cases of incorrect practices and cheating, such as keeping the poles off the ground and pole dragging. The \textsc{pamap2} data set\footnote{Available at \url{https://doi.org/10.24432/C5NW2H}, March 2024.} was used for further evaluation and comparison purposes. This data set contains data on various physical activities, including Nordic Walking, and was also used by Wiktorski \textit{et al.} (2019) \cite{Wiktorski2019}.

Due to the streaming implementation, two types of data were used in the different sub-problems considered: raw data and engineered features from raw data.

\begin{itemize}

 \item {\bf 1.} Raw data. Up to 54 features were considered at each slot for the \textsc{nwgti} data set, prior to feature engineering. These features directly corresponded to the sensor data available at the time (note that the sensors produce data at different rates, as described below). Forty features were extracted from the \textsc{pamap2} data set. Note that each individual had three wearable \textsc{imu}s: one on the chest and one each on the dominant wrist and dominant ankle. Each \textsc{imu} included a temperature sensor, two tri-axial accelerometers, a tri-axial gyroscope, a tri-axial magnetometer, and a heart rate sensor, which was only activated on the chest.

 \item {\bf 2.} Engineered features from raw data in the \textsc{nwgti} and \textsc{pamap2} data sets as described in the section \textsc{methodology. data calibration, feature engineering, analysis \& selection}.

\end{itemize}

\subsection*{WEARABLE SENSORS}
\label{sec:sensors_results}

As illustrated in Figure \ref{fig:sensors_position}, six wearable \textsc{imu}s by Mbientlab\footnote{Available at \url{https://www.bosch-sensortec.com/products/motion-sensors/imus/bmi160} and \url{https://mbientlab.com/store/metamotionrl}, March 2024.} were used to collect the data for the \textsc{nwgti} data. These \textsc{imu}s comprised tri-axial accelerometers, gyroscopes, and magnetometers, with respective sampling rates of \num{12.5}, \num{25}, and \num{10} Hz. Their maximum admissible acceleration was \num{16} g. Our experiments showed a maximum acceleration excess of 4.75 g, which is within a reasonable margin and indicates adequate sensitivity.

Note that the different \textsc{imu} sensors collected raw data at different rates. In principle thus, the assumption that all sensors are synchronized, implicit in the section \textsc{methodology. data calibration, feature engineering, analysis \& selection}, did not hold. The practical approach used to address this issue was to set ``empty'' $x_{p,a,s,l}[n]$ entries to NaN values at any input interrupt, which were ignored by feature computations (including online processing variability calculations).

Raw data from the \num{54} sensors available were produced during 12-minute average bursts. On average, each burst consisted of \num{94348} samples distributed as indicated in Table \ref{tab:dataset_distribution}. The number of feature dimensions depended on the data type (raw or engineered), as explained in the section \textsc{methodology. data calibration, feature engineering, analysis \& selection}.

\begin{table}[!htbp]
\centering
\caption{\label{tab:dataset_distribution}{Distribution of samples in the \textsc{nwgti} data set from data bursts (average values).}}
\begin{tabular}{lS[table-format=6.0]}
\toprule \textbf{Class} & \multicolumn{1}{c}{\textbf{Number of samples}}\\ \midrule
Correct practice & \num{30722} \\
Cheating practice & \num{33330}\\
Incorrect practice & \num{30296} \\
 \bottomrule
\end{tabular}
\end{table}

The sampling rate for the \textsc{pamap2} data set was \SI{100}{\hertz}, except for the heart rate sensor, which was \SI{9}{\hertz}. As previously mentioned, each \textsc{imu} in this case had two tri-axial accelerometers with respective maximum admissible accelerations of \SI{6}{g} and \SI{16}{g}. Thus, their sensitivities were also adequate for our application. Only Nordic walkers who walked for at least 250 seconds or users who climbed stairs for at least 150 seconds. This resulted in \num{43130} samples distributed as indicated in Table \ref{tab:dataset_distribution_pamap2}.

\begin{table}[!htbp]
\centering
\caption{\label{tab:dataset_distribution_pamap2}{Distribution of samples in the \textsc{pamap2} data set (average values).}}
\begin{tabular}{lS[table-format=6.0]}
\toprule \textbf{Class} & \multicolumn{1}{c}{\textbf{Number of samples}}\\ \midrule
Nordic Walking & \num{27974} \\
Climbing stairs & \num{15156} \\
 \bottomrule
\end{tabular}
\end{table}

\subsection*{SUB-PROBLEMS}
\label{subproblems}

For baseline data, we performed online processing supervised classifications with prequential evaluation by predicting, testing, and training the model in this specific order. In all cases, data were decimated for training and testing. Training and testing were updated every ten slots for the \textsc{nwgti} data set and, due to the higher sampling rate of \SI{100}{\hertz}, every 30 slots for the \textsc{pamap2} data set.

 \begin{itemize}
 \item \textbf{A}. Baseline supervised classification of the samples of experimental engineered data (in the sequel, by sample we will refer to all or part of the available features at the corresponding slot).
 
 \item \textbf{B}. Baseline shuffled supervised classification: the samples of experimental engineered data of sub-problem \textsc{a} were partitioned into eight randomly shuffled subsets. This emulates a situation in which a person performs correctly or incorrectly or cheats at different moments compared to sub-problem \textsc{a} and shows that there are no long-term dependencies in the data.
 
 \item \textbf{C}. Baseline stressed supervised classification: the samples in sub-problem \textsc{b} were decimated again for training and testing. Thus, training and testing were applied with the 1/100 ratio. This sub-problem stresses online processing supervised classification to check its robustness when reducing the proportion of annotated data.
 
 \item \textbf{D}. Unsupervised explainable classification: sample clusters from sub-problem \textsc{a} were used for unsupervised explainable re-classification.
\end{itemize}

\subsection*{DATA CALIBRATION, FEATURE ENGINEERING, ANALYSIS \& SELECTION}
\label{sec:data_processing_results}

\begin{description}

\item \textbf{Data calibration}. For the \textsc{nwgti} data set, the calibration method in the section \textsc{methodology. data calibration, feature engineering, analysis \& selection} yielded on average {$w_{avg}=362\pm58$, $w_{Q1}=168\pm0$, $w_{Q2}=282\pm6$ and $w_{Q3}=474\pm87$}, once the sensors had been synchronized with NaN padding. For the \textsc{pamap2} data set, {$w_{avg}=362\pm58$, $w_{Q1}=14\pm3$, $w_{Q2}=19\pm5$ and $w_{Q3}=25\pm5$}.

\item \textbf{Feature engineering}. In the \textsc{nwgti} data set, each engineered data sample had up to \num{24} features (the six features in Equation \eqref{eq:feature_engineering} for each of the four sliding windows) per sensor source, that is \num{24}$\times$\num{54}$=$\num{1296} features, plus up to \num{54} additional features directly corresponding to sensor outputs available at that moment. Therefore, each engineered data sample had up to \num{1350} features at each slot. For the \textsc{pamap2} data set, \num{24}$\times$\num{40}+40$=$\num{1000} features in total were used. These resulted from the 40 raw sensor measurements and the 24 engineered features for each of these measurements. Table \ref{tab:features} details the features engineered per \textsc{imu} for the two data sets.

\item \textbf{Feature analysis \& selection}. The \texttt{VarianceThreshold}\footnote{Available at \url{https://riverml.xyz/0.11.1/api/feature-selection/VarianceThreshold}, March 2024.} function from the \texttt{River}\footnote{Available at \url{https://riverml.xyz/0.11.1}, March 2024.} package was used to calculate online processed feature variances in the \textsc{nwgti} data set and select those exceeding $t^2_\sigma= 0.24\pm0.12$ in the raw data analysis and $t^2_\sigma=0.01\pm0.01$ in the engineered data scenario. This threshold was tuned as the median of the online processed variance of the features by considering the samples of a 60-second interval after the calibration stage. In the raw data analysis, at the last slot of the experiment, $27\pm4$ features were selected on average for sub-problems \textsc{a}, \textsc{b}, and \textsc{c}. The most relevant sensors were the accelerometers and the gyroscopes. In the case of engineered data, also at the last slot, $785\pm69$ features, of a total possible \num{1296} features were selected for sub problems \textsc{a} and \textsc{d}. The average number of features selected for sub-problems \textsc{b} and \textsc{c} was $789\pm59$. As in the case of raw data, the most relevant sensors were the accelerometers and the gyroscopes. The most relevant engineered feature was $F^w_{p,a,s,l}[n]$.

In the \textsc{pamap2} data set, the variance threshold was $t^2_\sigma= 8.99\pm3.28$ in the raw data analysis and $t^2_\sigma=5.86\pm1.58$ in the engineered data scenario. In the raw data analysis, on average, $23\pm2$ features were selected for sub problem \textsc{a} and $23\pm1$ for problems \textsc{b} and \textsc{c}. In the engineered data analysis, the average number of features selected was $526\pm32$ for sub problems \textsc{a} and \textsc{d} and $525\pm31$ for sub problems \textsc{b} and \textsc{c}. The most important features were those related to accelerometers, magnetometers, and heart rate sensors. As with the \textsc{nwgti} data set, the most relevant engineered feature was $F^w_{p,a,s,l}[n]$.

\end{description}

\begin{table*}[!htbp]
\centering
\footnotesize
\caption{\label{tab:features}Raw and engineered (Eng.) features per \textsc{imu} for the two data sets used.}
\begin{tabular}{ccp{5.2cm}c} 
\toprule
\bf Data set & \bf ID & \bf Name & \bf Type \\\midrule
\multirow{6}{*}{\textsc{nwgti}} & 1-3 & 16g accelerometer axes& \multirow{3}{*}{Raw}\\
& 4-6 & Gyroscope axes \\
& 7-9 & Magnetometer axes\\
&\multirow{3}{*}{10 - 225} & $Q^w_{1}, Q^w_{2}, Q^w_{3}, avg^w, std^w$ and $F^w$ for windows $w_{Q1}, w_{Q2}, w_{Q3}, w_{avg}$ for each sensor (1-9 above) & \multirow{3}{*}{Eng.}\\
\midrule
\multirow{7}{*}{\textsc{pamap2}} & 1-9 & Same as features 1-9 of \textsc{nwgti} data set & \multirow{3}{*}{Raw}\\
& 10-12 & 6g accelerometer axes \\
& 13 & Heart rate \\
& 14 & Temperature \\ 
&\multirow{3}{*}{15 - 350} & $Q^w_{1}, Q^w_{2}, Q^w_{3}, avg^w, std^w$ and $F^w$ for windows $w_{Q1}, w_{Q2}, w_{Q3}, w_{avg}$ for each sensor (1-14 above) & \multirow{3}{*}{Eng.}\\
\bottomrule
\end{tabular}
\end{table*}

\subsection*{ONLINE PROCESSING BASELINE AND UNSUPERVISED CLASSIFICATION}
\label{sec:classification_results}

The online processing supervised classification baseline algorithms were implemented with the \textsc{gnb}\footnote{Available at \url{https://riverml.xyz/dev/api/naive-bayes/GaussianNB}, March 2024.}, \textsc{hatc}\footnote{Available at \url{https://riverml.xyz/0.13.0/api/tree/HoeffdingAdaptiveTreeClassifier}, March 2024.}, \textsc{arfc}\footnote{Available at \url{https://riverml.xyz/dev/api/ensemble/AdaptiveRandomForestClassifier}, March 2024.}, and \textit{K}‐means\footnote{Available at \url{https://riverml.xyz/0.11.1/api/cluster/KMeans}, March 2024.} libraries from the \texttt{River} package.

Listings \ref{hatc_conf}, \ref{arfc_conf}, and \ref{kmeans_conf} respectively show the ranges for hyperparameter optimization (selected values in bold font) for the \textsc{hatc}, \textsc{arfc}, and {K}‐means models (the \textsc{gnb} model has no hyperparameters to tune). The ranges and best values were obtained experimentally using an \textit{ad hoc} implementation of the \texttt{\small GridSearch}\footnote{Available at \url{https://scikit-learn.org/stable/modules/generated/sklearn.model_selection.GridSearchCV.html}, March 2024.} algorithm for data streams. Engineered data for sub-problem \textsc{a} was down-sampled by a factor of \num{100} to perform hyperparameter optimization. The selected hyperparameters for both data sets were the same for \textsc{hatc} and \textsc{arfc}. For \textit{K}-means, however, the selected values for \texttt{halflife} (which is related to centroid optimization) varied depending on the data set. For \textsc{pamap2}, the selected value within the specified sequence was 0.77 (not shown). Recall that the goal with this model was to discover three clusters for \textsc{nwgti} and two for \textsc{pamap2} (\textit{i.e.}, \texttt{n\_clusters} = 3 or 2). 

\begin{lstlisting}[frame=single,float,caption={\textsc{hatc} hyperparameter ranges and selected values in bold.},label={hatc_conf},emphstyle=\textbf,escapechar=ä]
depth = [ä\textbf{50}ä, 100, 150]
tiethreshold = [0.5, ä\textbf{0.05}ä, 0.005]
maxsize = [ä\textbf{50}ä, 100, 200]
\end{lstlisting}

\begin{lstlisting}[frame=single,float=htb,caption={\textsc{arfc} hyperparameter ranges and selected values in bold.},label={arfc_conf},emphstyle=\textbf,escapechar=ä]
models = [ä\textbf{50}ä, 100, 200]
features = [ä\textbf{50}ä, 100, 200]
lambda = [ä\textbf{50}ä, 100, 200]
\end{lstlisting}

\begin{lstlisting}[frame=single,float,caption={\textit{K}‐means hyperparameter ranges and selected values in bold.},label={kmeans_conf},emphstyle=\textbf,escapechar=ä]
halflifeä$_{nwgti}$ä = [0.05, ä\textbf{0.075}ä, 0.1]
halflifeä$_{pamap2}$ä = [0.02:0.05:0.8]
mu = [ä\textbf{0.01}ä, 0.1, 1]
sigma = [ä\textbf{0.001}$_{nwgti}$ä, 1, ä\textbf{10}$_{pamap2}$ä]
p = [ä\textbf{1}ä, 2]
\end{lstlisting}

Table \ref{tab:classification_results} reports standard performance measurements by type of experimental data, sub-problem, and model (tags \#1, \#2, and \#3 correspond to correct, cheating, and incorrect practices, respectively) for the \textsc{nwgti} data set. It allows comparison between our unsupervised solution (scenario \textsc{d} with engineered features) and the supervised learning baselines (scenarios \textsc{a}, \textsc{b}, and \textsc{c}, for both raw data and engineered features). All entries average the corresponding results for the user sessions in the experiments. Column ``prequential time'' is the total time needed to process all the samples for each model and sub-problem. It should be checked against the maximum data rate of the sensors, 25 Hz. In other words, for a method to be feasible in real-time, its prequential time per sample must be less than 40 ms (in our worst-case scenario, it is 3 ms, that is \SI{306.35}{\second} divided by \num{94348} samples, about \SI{7.5}{\percent} of the limit value). It should be noted that all methods except $K$-means are used for obtaining baseline data due to their supervised nature, which is impractical for real use. Thus, in light of the results obtained, $K$-means is feasible.

First, it is interesting to observe that sub-problem \textsc{a} was highly separable when only raw data were used. The slowest method was \textsc{arfc}, but it was also feasible in streaming mode. As expected, the results of sub-problem \textsc{b} suggest no long-term dependence. For sub-problem \textsc{c}, the outcome was considerably degraded, suggesting that, although the approaches seem robust, excessive decimation is not conducive to real-time operation on simple hardware since it compromises performance.

The performance gap between the different methods was reduced by introducing engineered data. The most relevant result here, highlighting the value of engineered data, was the significant improvement in sub-problem \textsc{c}, regardless of its low computational requirements. For this reason, we decided to test the practical, unsupervised approach of sub-problem \textsc{d} using engineered data. As shown in the last row of Table \ref{tab:classification_results}, performance was highly satisfactory, assuming each cluster is assigned to the correct class {\it a posteriori} to ensure minimum tagging. It should be noted that, in a competition scenario, it would not be necessary to determine which of the two differentiated incorrect behaviors corresponds to cheating and which to incorrect practice, since the correct practice cluster could be identified from the few points along the route where the practitioner passes by a judge. Moreover, a coach could differentiate them using training videos {\it a posteriori}.

After sub-problem \textsc{d}, with the inclusion of re-classification for explainability, the prequential processing time is the sum of the corresponding successive times of \textit{K}-means and \textsc{arfc}, that is, 224.15 ms, which is still feasible for real-time operation.

\begin{table*}[!htbp]
\scriptsize
\centering
\caption{\label{tab:classification_results}Online processing practice assessment results, \textsc{nwgti} data set.}
\begin{tabular}{ccccccccccccS[table-format=3.2]}
\toprule
\bf Data & \bf Sce. & \bf Model & \bf Accuracy & \multicolumn{4}{c}{\bf Precision} & \multicolumn{4}{c}{\bf Recall} & {\bf Preq. time (s)}\\
\cmidrule(lr){5-8}
\cmidrule(lr){9-12}
 & & & & Macro & \#1 & \#2 & \#3 & Macro & \#1 & \#2 & \#3 \\
\midrule
\multirow{9}{*}{Raw} & 
\multirow{3}{*}{A} & 
\textsc{gnb} & $90.99 \pm7.04$ & 91.86 & 93.27 & 93.38 & 88.93 & 91.15 & 94.45 & 90.30 & 88.71 & 1.79\\

& & 
\textsc{hatc} & $96.94 \pm1.80$ & 97.19 & 97.90 & 95.75 & \bf 97.92 & 96.90 & 98.38 & \bf 98.84 & 93.47 & 3.07\\
& & 
\textsc{arfc} & $\bf99.03 \pm\bf0.55$ & \bf 99.03 & \bf 99.83 & \bf 99.65 & 97.60 & \bf 99.02 & \bf 99.28 & 98.23 & \bf 99.55 & 164.68\\

\cmidrule(lr){2-13}

& \multirow{3}{*}{B} & 
\textsc{gnb} & $79.87 \pm0.71$ & 80.39 & 79.49 & 82.73 & 78.95 & 79.43 & 71.19 & 87.88 & 79.22 & 2.05\\
& & 
\textsc{hatc} & $\bf92.98 \pm\bf2.85$ & \bf 92.95 & \bf 91.85 & \bf 93.59 & \bf 93.42 & \bf 93.05 & \bf 94.30 & \bf 93.19 & \bf 91.65 & 3.83\\
& & 
\textsc{arfc} & $88.77 \pm4.39$ & 89.00 & 83.38 & 91.45 & 92.16 & 88.76 & 91.97 & 91.06 & 83.25 & 306.35\\

\cmidrule(lr){2-13}

& \multirow{3}{*}{C} & 
\textsc{gnb} & $\bf75.80 \pm\bf2.31$ & \bf 76.14 & \bf 73.28 & \bf 82.31 & 72.84 & \bf 75.26 & 65.49 & \bf 84.75 & \bf 75.53 & 1.87\\
& & 
\textsc{hatc} & $71.92 \pm11.63$ & 71.52 & 67.08 & 74.89 & 72.57 & 71.25 & 64.50 & 83.25 & 66.00 & 1.50\\
& & 
\textsc{arfc} & $70.46 \pm9.16$ & 71.25 & 65.62 & 72.31 & \bf75.82 & 70.24 & \bf70.82 & 77.90 & 61.98 & 139.46\\
\midrule

\multirow{10}{*}{Eng.} & 
\multirow{3}{*}{A} & 
\textsc{gnb} & $95.39 \pm2.35$ & 95.55 & 98.15 & 96.36 & 92.15 & 95.42 & 93.65 & 95.67 & 96.94 & 38.47\\
& & 
\textsc{hatc} & $93.53 \pm7.13$ & 94.64 & 89.71 & 97.48 & 96.74 & 93.53 & 99.36 & 98.02 & 83.21 & 58.55\\
& & 
\textsc{arfc} & $\bf99.36 \pm\bf0.31$ & \bf 99.38 & \bf99.59 & \bf99.21 & \bf99.33 & \bf 99.36 & \bf99.91 & \bf99.44 & \bf98.74 & 178.02\\

\cmidrule(lr){2-13}

& \multirow{3}{*}{B} & 
\textsc{gnb} & $93.25 \pm4.55$ & 93.40 & 89.79 & 96.23 & 94.20 & 93.13 & 95.17 & 94.03 & 90.20 & 42.56\\
& & 
\textsc{hatc} & $92.17 \pm2.64$ & 92.34 & 91.87 & 94.46 & 90.68 & 92.21 & 93.93 & 93.10 & 89.61 & 83.36\\
& & 
\textsc{arfc} & $\bf98.79 \pm\bf0.90$ & \bf 98.78 & \bf98.88 & \bf99.20 & \bf98.25 & \bf 98.75 & \bf97.87 & \bf99.45 & \bf98.92 & 281.89\\

\cmidrule(lr){2-13}

& \multirow{3}{*}{C} & 
\textsc{gnb} & $92.44 \pm4.77$ & 92.70 & 88.40 & 95.48 & 94.23 & 92.31 & 93.46 & 93.46 & 90.01 & 36.69\\
& & 
\textsc{hatc} & $91.05 \pm5.22$ & 91.32 & 90.56 & 91.66 & 91.75 & 90.89 & 91.63 & 93.02 & 88.01 & 45.36\\
& & 
\textsc{arfc} & $\bf97.63 \pm\bf0.66$ & \bf 97.58 & \bf97.12 & \bf98.52 & \bf97.09 & \bf 97.61 & \bf97.32 & \bf97.54 & \bf97.97 & 117.57\\

\cmidrule(lr){2-13}

& D & Clustering & $\bf97.68 \pm\bf0.83$ & \bf 97.70 & \bf98.37 & \bf98.61 & \bf96.14 & \bf 97.74 & \bf98.60 & \bf95.84 & \bf98.78 & 46.13\\

\bottomrule
\end{tabular}
\end{table*}

Table \ref{tab:classification_results_pamap2} shows the results obtained with the \textsc{pamap2} data set. Although this data set is designed for activity differentiation not practice assessment (tags \#1 and \#2 correspond to Nordic Walking and stair climbing, respectively), the good performance achieved with our solution highlights the good quality of the data collected in \textsc{nwgti}. The unsupervised model attained performance metrics above \SI{90}{\percent} for Nordic Walking prediction, although the \textsc{gnb} classifier was the best model in this case. Elapsed times were considerably reduced due to the fewer samples and classes.

\begin{table*}[!htbp]
\scriptsize
\centering
\caption{\label{tab:classification_results_pamap2}Online processing practice assessment results, \textsc{pamap2} data set.}
\begin{tabular}{ccccccccccS[table-format=3.2]}
\toprule
\bf Data & \bf Sce. & \bf Model & \bf Accuracy & \multicolumn{3}{c}{\bf Precision} & \multicolumn{3}{c}{\bf Recall} & {\bf Preq. time (s)}\\
\cmidrule(lr){5-7}
\cmidrule(lr){8-10}
 & & & & Macro & \#1 & \#2 & Macro & \#1 & \#2 \\
\midrule
\multirow{10}{*}{Raw} & 
\multirow{3}{*}{A} & 
\textsc{gnb} & $95.69 \pm4.94$ & 95.23 & 100.00 & 90.46 & 96.57 & 93.14 & 100.00 & 0.21\\
& & 
\textsc{hatc} & $82.74 \pm16.80$ & 86.47 & 100.00 & 72.94 & 86.49 & 72.98 & 100.00 & 0.32\\
& & 
\textsc{arfc} & $\bf98.70 \pm\bf0.67$ & \bf98.25 & \bf100.00 & \bf96.49 & \bf98.98 & \bf97.96 & \bf100.00 & 16.90\\

\cmidrule(lr){2-11}

& \multirow{3}{*}{B} & 
\textsc{gnb} & $\bf93.13 \pm\bf4.71$ & \bf92.33 & \bf95.98 & \bf\bf88.67 & \bf93.15 & 93.09 & \bf93.21 & 0.22\\
& & 
\textsc{hatc} & $86.59 \pm6.32$ & 85.96 & 87.63 & 84.28 & 83.89 & 92.76 & 75.01 & 0.41\\
& & 
\textsc{arfc} & $87.08 \pm2.47$ & 88.21 & 85.68 & 90.74 & 83.24 & \bf96.23 & 70.25 & 24.47\\

\cmidrule(lr){2-11}

& \multirow{3}{*}{C} & 
\textsc{gnb} & $\bf90.33 \pm\bf4.77$ & \bf89.76 & \bf91.84 & \bf87.69 & \bf89.10 & 93.17 & \bf85.03 & 0.19\\
& & 
\textsc{hatc} & $80.04 \pm12.82$ & 83.61 & 80.81 & 86.41 & 74.93 & \bf94.79 & 55.08 & 0.16\\
& & 
\textsc{arfc} & $70.97 \pm6.65$ & 70.08 & 72.27 & 67.89 & 63.18 & 89.96 & 36.40 & 10.44\\
\midrule

\multirow{10}{*}{Eng.} & 
\multirow{3}{*}{A} & 
\textsc{gnb} & $\bf99.00 \pm\bf0.21$ & \bf98.62 & \bf100.00 & \bf97.24 & \bf99.23 & \bf98.46 & \bf100.00 & 3.61\\
& & 
\textsc{hatc} & $91.93 \pm10.43$ & 92.35 & 100.00 & 84.69 & 93.85 & 87.70 & 100.00 & 6.02\\
& & 
\textsc{arfc} & $96.99 \pm0.72$ & 95.98 & 100.00 & 91.97 & 97.69 & 95.38 & 100.00 & 35.11\\

\cmidrule(lr){2-11}

& \multirow{3}{*}{B} & 
\textsc{gnb} & $\bf99.27 \pm\bf0.28$ & \bf99.26 & \bf99.33 & \bf99.18 & \bf99.15 & \bf99.55 & \bf98.75 & 3.79\\
& & 
\textsc{hatc} & $95.87 \pm3.40$ & 95.58 & 96.58 & 94.57 & 95.38 & 97.27 & 93.50 & 7.60\\
& & 
\textsc{arfc} & $96.18 \pm2.71$ & 96.00 & 96.62 & 95.38 & 95.64 & 97.51 & 93.77 & 66.65\\
\cmidrule(lr){2-11}

& \multirow{3}{*}{C} & 
\textsc{gnb} & $\bf97.25 \pm\bf0.60$ & \bf97.42 & 97.00 & \bf97.83 & \bf96.54 & \bf98.81 & 94.27 & 3.17\\
& & 
\textsc{hatc} & $89.12 \pm11.57$ & 92.50 & 87.32 & 97.68 & 85.40 & 99.35 & 71.45 & 3.21\\
& & 
\textsc{arfc} & $96.59 \pm2.57$ & 96.38 & \bf97.26 & 95.51 & 96.13 & 97.54 & \bf94.72 & 22.12\\

\cmidrule(lr){2-11}

& D & Clustering & $\bf90.46 \pm\bf11.39$ & \bf93.57 & \bf92.26 & \bf94.88 & \bf87.97 & \bf95.53 & \bf80.41 & 3.40\\

\bottomrule
\end{tabular}
\end{table*}

\begin{figure}[!htbp]
\centering
\includegraphics[width=0.5\textwidth]{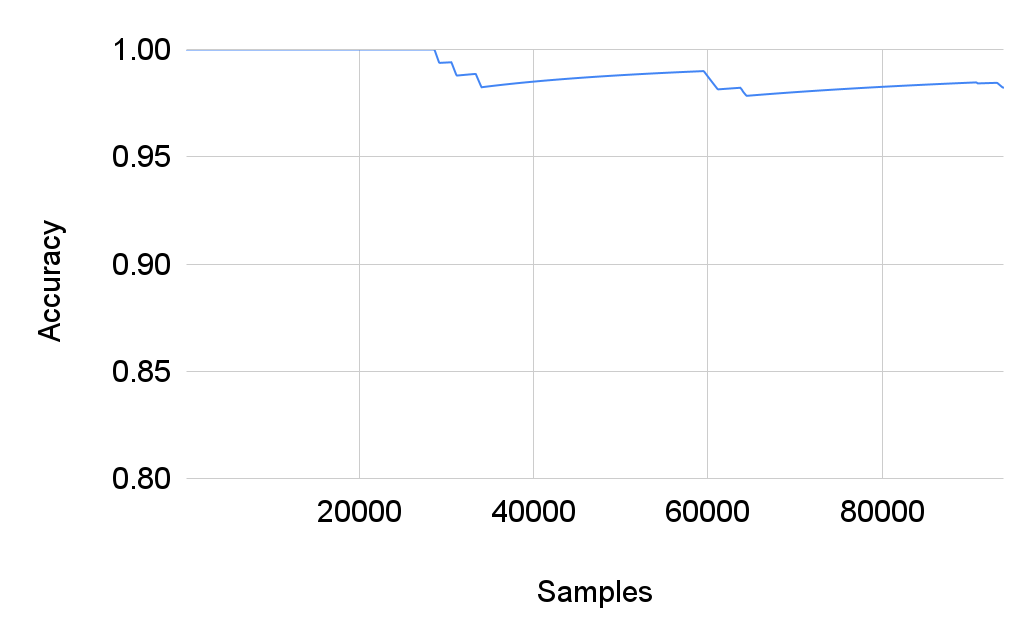}
\caption{\label{fig:acc_our}Accuracy curve for the \textsc{nwgti} data set.}
\end{figure}

\begin{figure}[!htbp]
\centering
\includegraphics[width=0.5\textwidth]{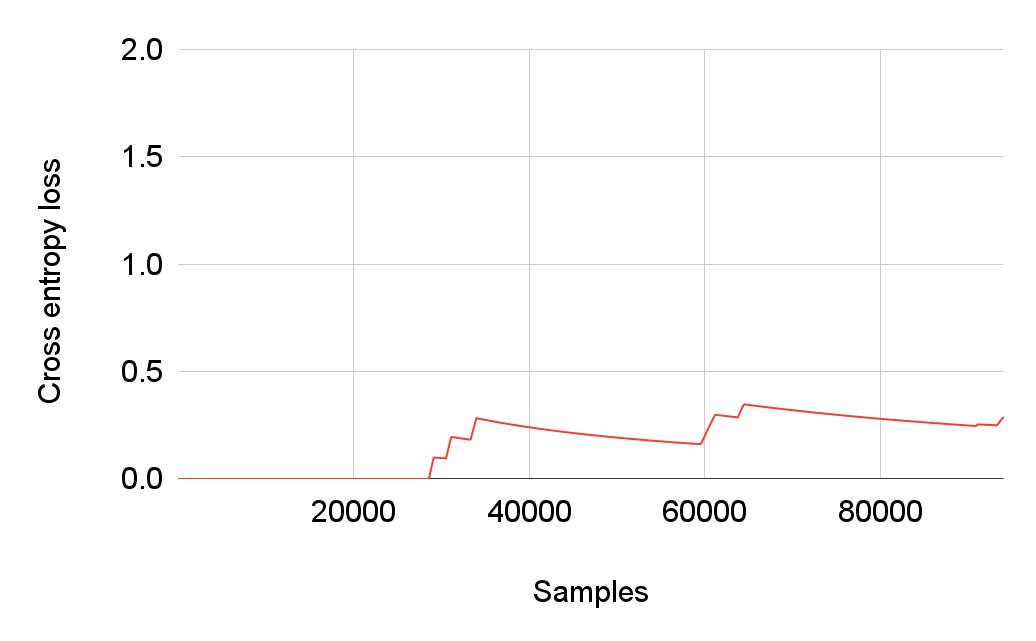}
\caption{\label{fig:error_our}Cross entropy loss curve for the \textsc{nwgti} data set.}
\end{figure}

Figures \ref{fig:acc_our} and \ref{fig:error_our} respectively show online changes in accuracy and cross entropy loss curves for scenario \textsc{d} (unsupervised learning) with the \textsc{nwgti} data set. The accuracy was slightly reduced in the last two-thirds of the sequence, but it was still very good. To calculate the cross entropy loss graph we employed Equation \eqref{eq:cross_entropy}, where $E$ is the number of samples, $M$ the number of classes, and $h$ a one-hot encoded vector (in our experiment, as there exist three classes, if the sample corresponds to the second class, the vector was (0,1,0)), and $\rho_i[n]$ is the probability of predicting class $i$ for the $n$-th sample). The graph showed a slight increase in accumulated error, but note that values lower than 0.5 are satisfactory given the logarithmic term in the $crloss$ function.

\begin{equation}\label{eq:cross_entropy}
\begin{split}
crloss=-\frac{1}{E}\sum_{i}^{E}\sum_{j}^{M} h_{j}[i]log(\rho_{j}[i])
\end{split}
\end{equation}

Finally, Table \ref{tab:comparison_results} compares our approach with the most closely related work in the literature. The authors, Derungs \textit{et al.} (2018) \cite{Derungs2018}, computed the same evaluation metrics in that work were computed, \textsc{rmse} and \textsc{mae}. The lower error observed with our system can be explained by the fact that Derungs \textit{et al.} (2018) \cite{Derungs2018} only considered classifiers as regressors and did not use nominal categories, which we used to explore strategies such as the \textsc{arfc} model. Furthermore, they grouped sensor data into individual strides, preventing continuous signal analysis. Finally, this comparison must be understood as a reference since the approach in Derungs \textit{et al.} \cite{Derungs2018} was supervised, and our premise was to avoid that limitation to the greatest possible extent.

\begin{table}[!htbp]
\centering
\caption{\label{tab:comparison_results}{Comparison with the most related work.}}
\begin{tabular}{lcc}
\toprule \textbf{Approach} & \bf \textsc{rmse} & \bf \textsc{mae}\\ \midrule
This work on \textsc{nwgti} & 0.18 & 0.03\\
Derungs \textit{et al.} (2018) \cite{Derungs2018} & 0.43 & 0.18 \\
 \bottomrule
\end{tabular}
\end{table}

\subsection*{UNSUPERVISED EXPLAINABILITY}
\label{sec:explainability_results}

\begin{figure*}[!htbp]
\centering
\includegraphics[width=1\textwidth]{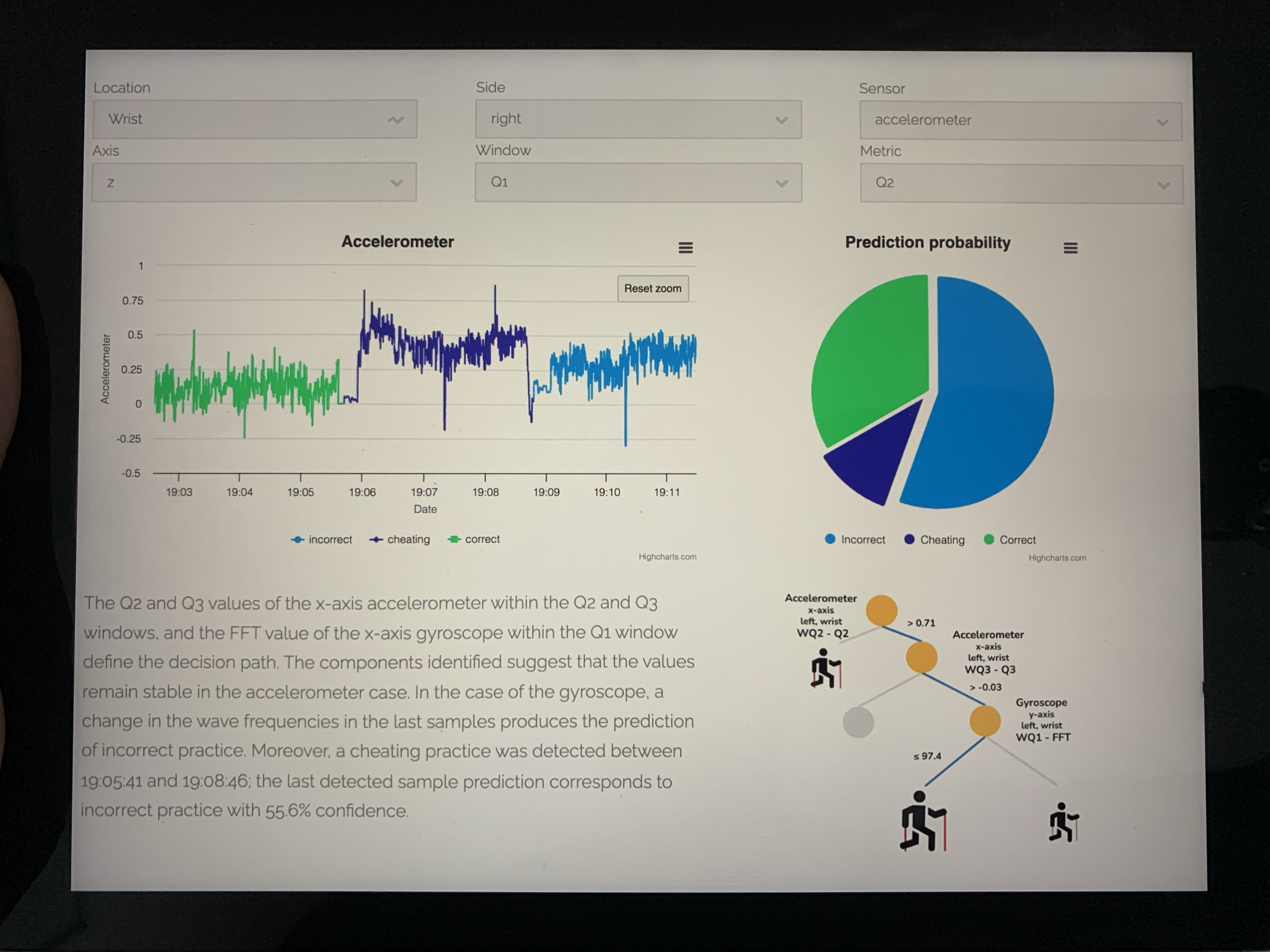}
\caption{\label{fig:dashboard}Screenshot of explainability dashboard.}
\end{figure*}

The \textsc{arfc} algorithm is the basis for explainability once trained with the clusters returned by the online processing unsupervised \textit{K}-means algorithm. It is based on the Hoeffding tree model \cite{Gomes2017}. The explainability module groups the trees from the \textsc{arfc} model by predicted categories. Then, the \texttt{River} \texttt{debug\_one} \textsc{arfc} method\footnote{Available at \url{https://riverml.xyz/0.14.0/api/tree/HoeffdingTreeClassifier}, March 2024.} is used to extract the decision paths traversed, whose features are stored in lists. The lists are then combined, and the frequency with which feature components appear is calculated as explained in the section \textsc{explainable unsupervised re-classification}. At this point, the feature components are ranked in decreasing order of relevance. In Figure \ref{fig:dashboard}, the most representative feature component to that point, corresponding to the $z$-axis of the right wrist accelerometer and the $w_{Q1}$ sliding window of the $Q^w_2$ component, is displayed. Other feature components can be selected for display using the drop box on the dashboard.

The dashboard was designed with \textsc{templated}\footnote{Available at \url{https://templated.co}, March 2024.} and its charts with Highcharts\footnote{Available at \url{https://www.highcharts.com}, March 2024.}. Figure \ref{fig:dashboard} shows an automatic explanation of a training session of one of the Nordic Walking practitioners in our experiments.

The upper part shows the selected sensor as previously defined. The real-time graph for the sensor is shown on the left, where green, purple, and blue identify correct, cheating, and incorrect practices, respectively. The description of the latest prediction in natural language, based on the \textsc{arfc} decision path, is the text in the bottom left corner, corresponding to the tree path displayed in the bottom right corner. The extracted information is introduced into natural language templates taken from Listing \ref{templates_conf} (where ellipses indicate that the immediately preceding text is repeated as needed). Our system follows the tree path of the latest prediction and extracts all associated index tuples $(a,s)$. In case of repetition, an index tuple $(a,s)$ is only considered once. For each instance, template \#1 is updated with the lists of different components and windows across the decision path corresponding to the selected tuples. Template \#2 is applied to all nodes along the decision path in which the '$\leq$' condition holds for a component whose updated standard deviation is less than $t_\sigma$. For each node along the tree path, template \#3 is updated whenever a '$>$' condition holds for the sensor and component considered in that node. Finally, template \#4 identifies the cheating interval, if present, and the current prediction, including its confidence interval.

\begin{lstlisting}[frame=single,float,caption={Explainability templates.},label={templates_conf},emphstyle=\textbf,escapechar=ä]
#1 The [ä\textcolor{new_blue}{\textbf{component list}}ä] valueä\textcolor{new_blue}{\textbf{<s>}}ä of the 
[ä\textcolor{new_blue}{\textbf{axis}}ä] [ä\textcolor{new_blue}{\textbf{sensor}}ä] within the [ä\textcolor{new_blue}{\textbf{windows list}}ä] 
windowä\textcolor{new_blue}{\textbf{<s>}}ä and ... define the decision
path. 

#2 The componentä\textcolor{new_blue}{\textbf{<s>}}ä identified 
suggest that the valueä\textcolor{new_blue}{\textbf{<s>}}ä remain 
stable in the [ä\textcolor{new_blue}{\textbf{sensor}}ä] case.

#3 In the case of the [ä\textcolor{new_blue}{\textbf{sensor}}ä] a 
change in the [ä\textcolor{new_blue}{\textbf{component}}ä] in the last 
samples and ... produceä\textcolor{new_blue}{\textbf{<s>}}ä the 
prediction of [ä\textcolor{new_blue}{\textbf{correct}}ä/ä\textcolor{new_blue}{\textbf{cheating}}ä/ä\textcolor{new_blue}{\textbf{incorrect}}ä]
practice. 

#4 Moreover, a cheating practice was 
detected between [ä\textcolor{new_blue}{\textbf{start time}}ä] and 
[ä\textcolor{new_blue}{\textbf{end time}}ä]; the last detected sample 
prediction corresponds to [ä\textcolor{new_blue}{\textbf{correct}}ä/
ä\textcolor{new_blue}{\textbf{cheating}}ä/ä\textcolor{new_blue}{\textbf{incorrect}}ä] practice with 
[ä\textcolor{new_blue}{\textbf{confidence}}ä] confidence.

\end{lstlisting}

The $w_{Q1}$, $w_{Q2}$, $w_{Q3}$, and $w_{avg}$ windows determine the time duration under analysis. When the features in the decision path displayed in the dashboard correspond to the $w_{Q1}$ window (as in the example), the value changes are recent. Features corresponding to the $w_{Q3}$ window, by contrast, are more spaced in time. The $Q^w_{1}$, $Q^w_{2}$, $Q^w_{3}$, $avg^w$, $std^w$, and $F^w$ features allow for the interpretation of extreme values. Specifically, a value in $Q^w_{1}$ would indicate that only \SI{25}{\percent} of the samples have lower values. $F^w$ indicates a change in the trend of the movement. The prediction confidence was obtained using the \textsc{arfc} \texttt{predict\_proba\_one} method. This information is displayed graphically in the tree in the bottom right of the image in Figure \ref{fig:dashboard}.

\section*{CONCLUSIONS}
\label{sec:conclusions}

Evaluation of sport and physical activity, including aspects such as performance monitoring and injury prevention, is a major field of application for \textsc{ai} \textsc{har} techniques. \textsc{ml}-based solutions are helpful for differentiating between well-defined states within large volumes of data collected using inexpensive portable sensors such as wearables and smartphones. Unsupervised techniques are very interesting in this context as they can be used in the design of stand-alone systems. Nordic Walking \textsc{har}, the core application of this research, aligns well with these assumptions, and can be used to differentiate between correct and incorrect techniques and practices that would result in disqualification. This information would be useful for practitioners, trainers, and competition judges. 

In the proposed solution, supervised (baseline) and unsupervised models were trained using data from wearable inertial sensors. These data were then augmented using data engineering techniques to allow long processing slots. Experimental testing demonstrated the appropriateness of the unsupervised learning approach, as it achieved a classification accuracy of close to \SI{100}{\percent} (note that the experimental evaluation also considered an alternative data set from the literature and showed that our data collection was not biased toward improving performance results). The outcome of the unsupervised learning stage (\textit{i.e.}, the resulting clusters) yields labels at minimal cost for automatic explainable re-classification. By reducing thus the need for laborious manual tagging, without compromising interpretability, our work contributes to solving a common need in \textsc{har} for consumer electronics applications.

The automatic explanations generated for the classification results include textual and visual descriptions of athletes' performance as well as intelligible explanations of the predictions made by the \textsc{ml} models. The ultimate objective is to promote trust, transparency, and acceptance for \textsc{ai} solutions. 

Automatic explainability has received very little attention in sports to date, with solutions focusing on offline supervised approaches. To our knowledge, our system is the first solution to combine online processing, explainability, and unsupervised assessment for \textsc{har} in the field of sport. The flexible positioning of the data collection sensors is an additional advantage of our approach. 

Currently, use of our system is limited to sports or training practices with clearly defined patterns or stages, such as martial arts kata training, boxing bag exercises, certain forms of gymnastics (\textit{e.g.}, rings), and, as shown in this paper, Nordic Walking. Future work is needed to extend its application to more open, ``fluid'', sports, such as soccer and basketball. In addition, when applicable, we plan to take advantage of the solution's modular design to include a reinforcement learning module based on the \textit{human in the loop} approach. This would allow experts to provide feedback to continually improve the system and its performance.

\subsection*{ACKNOWLEDGMENTS}

This work was partially supported by Xunta de Galicia grants ED481B-2021-118, ED481B-2022-093, and ED431C 2022/04, Spain. The authors are indebted to Nordic Walking Vigo and Mr. Ignacio Garc\'ia P\'erez for their help obtaining representative experimental data for correct, incorrect, and cheating practices.

\bibliographystyle{IEEEtran}
\bibliography{3_mybibfile}

\begin{IEEEbiography}{Silvia García-Méndez} received a Ph.D. in Information and Communication Technologies from the University of Vigo in 2021. Since 2015, she has worked as a researcher with the Information Technologies Group at the University of Vigo. She is collaborating with foreign research centers as part of her postdoctoral stage. Her research interests include Natural Language Processing techniques and Machine Learning algorithms.
\end{IEEEbiography}

\begin{IEEEbiography}{Francisco de Arriba-Pérez} received a B.S. degree in telecommunication technologies engineering in 2013, an M.S. degree in telecommunication engineering in 2014, and a Ph.D. degree in 2019 from the University of Vigo, Spain. He is currently a researcher in the Information Technologies Group at the University of Vigo, Spain. His research includes the development of Machine Learning solutions for different domains like finance and health.
\end{IEEEbiography}

\begin{IEEEbiography}{Francisco J. González-Castaño} received a B.S. degree from the University of Santiago de Compostela, Spain, in 1990 and a Ph.D. degree from the University of Vigo, Spain, in 1998. He is a full professor at the University of Vigo, Spain, leading the Information Technologies Group. He has authored over 120 papers in international journals in the ﬁelds of telecommunications and computer science and has participated in several relevant national and international projects. He holds three U.S. patents.
\end{IEEEbiography}

\begin{IEEEbiography}{Javier Vales-Alonso} received a degree in
telecommunication engineering from the Universidad de Vigo, Spain, in 2000, the M.Sc. degree in mathematics from the Universidad Nacional de Educación a Distancia, Spain, in 2005, and the Ph.D. degree in computer science from the Universidad Politécnica de Cartagena (UPCT), Spain, in 2015, where he is currently a Full Professor with the Department of Information and Communication Technologies. He is involved in different research topics related to modeling and optimization.
\end{IEEEbiography}

\end{document}